\documentclass[a4paper,fleqn]{cas-dc}

\usepackage[numbers]{natbib}
\usepackage{subfigure}
\usepackage{algorithm}
\usepackage{algorithmic}
\usepackage{amsmath,amsfonts,amsthm}
\usepackage{amssymb}             
\usepackage{mathrsfs}            
\usepackage{paralist}
\usepackage{multirow,multicol}
\usepackage{tabularx}
\usepackage{booktabs}
\usepackage{subfigure}
\usepackage{colortbl}
\usepackage{makecell}
\usepackage{color}
\usepackage{array}
\usepackage{mdwmath}
\usepackage{mdwtab}
\usepackage{eqparbox}
\usepackage{url}
\usepackage{microtype}
\usepackage{adjustbox}
\usepackage{listings}
\usepackage{paralist}
\lstdefinestyle{Python}{
	language        =   Python, 
	basicstyle      =   \zihao{-5}\ttfamily,
	numberstyle     =   \zihao{-5}\ttfamily,
	keywordstyle    =   \color{blue},
	keywordstyle    =   [2] \color{teal},
	stringstyle     =   \color{magenta},
	commentstyle    =   \color{red}\ttfamily,
	breaklines      =   true,   
	columns         =   fixed,  
	basewidth       =   0.5em,
}
\definecolor{Blue2}{RGB}{86, 160, 255}

\def\tsc#1{\csdef{#1}{\textsc{\lowercase{#1}}\xspace}}
\tsc{WGM}
\tsc{QE}
\tsc{EP}
\tsc{PMS}
\tsc{BEC}
\tsc{DE}

\begin{document}
\let\WriteBookmarks\relax
\def\floatpagepagefraction{1}
\def\textpagefraction{.001}
\shorttitle{CTISum}
\shortauthors{Wei Peng et~al.}

\title [mode = title]{CTISum: A New Benchmark Dataset For Cyber Threat Intelligence Summarization}                      
\tnotemark[1]

\tnotetext[1]{This document is the results of the research
project funded by the Zhongguancun Laboratory.}

\author[1]{Wei Peng}[ orcid=0000-0001-8179-1577]
\ead{pengwei@zgclab.edu.cn}

\author[2]{Junmei Ding}[style=chinese]

\author[1]{Wei Wang}
\ead{wangwei@zgclab.edu.cn}

\author[1]{Lei Cui}
\ead{cuilei@zgclab.edu.cn}

\author[1]{Wei Cai}
\ead{caiwei@zgclab.edu.cn}

\author[1]{Zhiyu Hao}
\ead{haozy@zgclab.edu.cn}

\author[1]{Xiaochun Yun}
\ead{yunxiaochun@zgclab.edu.cn}

\address[1]{Zhongguancun Laboratory, Beijing, P.R. China}

\address[2]{Beijing University of Posts and Telecommunications, Beijing, China}

\cortext[cor1]{Corresponding author}

\begin{abstract}
Cyber Threat Intelligence (CTI) summarization involves generating concise and accurate highlights from web intelligence data, which is critical for providing decision-makers with actionable insights to swiftly detect and respond to cyber threats in the cybersecurity domain. Despite that, the development of efficient techniques for summarizing CTI reports, comprising facts, analytical insights, attack processes, and more, has been hindered by the lack of suitable datasets. To address this gap, we introduce CTISum, a new benchmark dataset designed for the CTI summarization task. Recognizing the significance of understanding attack processes, we also propose a novel fine-grained subtask: attack process summarization, which aims to help defenders assess risks, identify security gaps, and uncover vulnerabilities. Specifically, a multi-stage annotation pipeline is designed to collect and annotate CTI data from diverse web sources, alongside a comprehensive benchmarking of CTISum using both extractive, abstractive and LLMs-based summarization methods. Experimental results reveal that current state-of-the-art models face significant challenges when applied to CTISum, highlighting that automatic summarization of CTI reports remains an open research problem. The code and example dataset can be made publicly available at https://github.com/pengwei-iie/CTISum.
\end{abstract}


\begin{highlights}
\item To the best of our knowledge, we make the first attempt to build a new benchmark CTISum with the CTIS task and a novel APS subtask in the cybersecurity domain.
\item A multi-stage annotation pipeline is designed to obtain the high-quality dataset with the assistance of LLMs while manually controlling the quality.
\item Experiments on CTISum demonstrate the challenge of the proposed two tasks, meanwhile indicating a large space for future research.
\end{highlights}


\maketitle

\section{Introduction}
Cyber Threat Intelligence (CTI), also known as threat intelligence, is knowledge, skills and experience-based information concerning the occurrence and assessment of both cyber and physical threats as well as threat actors \cite{DBLP:journals/compsec/JoLS22,bank2016cbest}. 
The CTI data often originates from diverse web sources, including forums, blogs, and open-source repositories, making it difficult for analysts to efficiently locate the most relevant and high-value intelligence from the web. 
Therefore, it becomes crucial to automatically summarize the knowledge contained in CTI reports, 
which could help analysts simply identify events, patterns, cyber attacks and conclusions. 
Furthermore, CTI summarization has broad applicability across domains like cybersecurity \cite{DBLP:journals/jdfsl/SchatzBW17}, military intelligence \cite{DBLP:reference/crypt/2005}, technical alerts, threat modeling, and more, where complicated information needs to be distilled into concise insights.

\begin{figure}[t]
	\centering
	\includegraphics[width=0.49\textwidth]{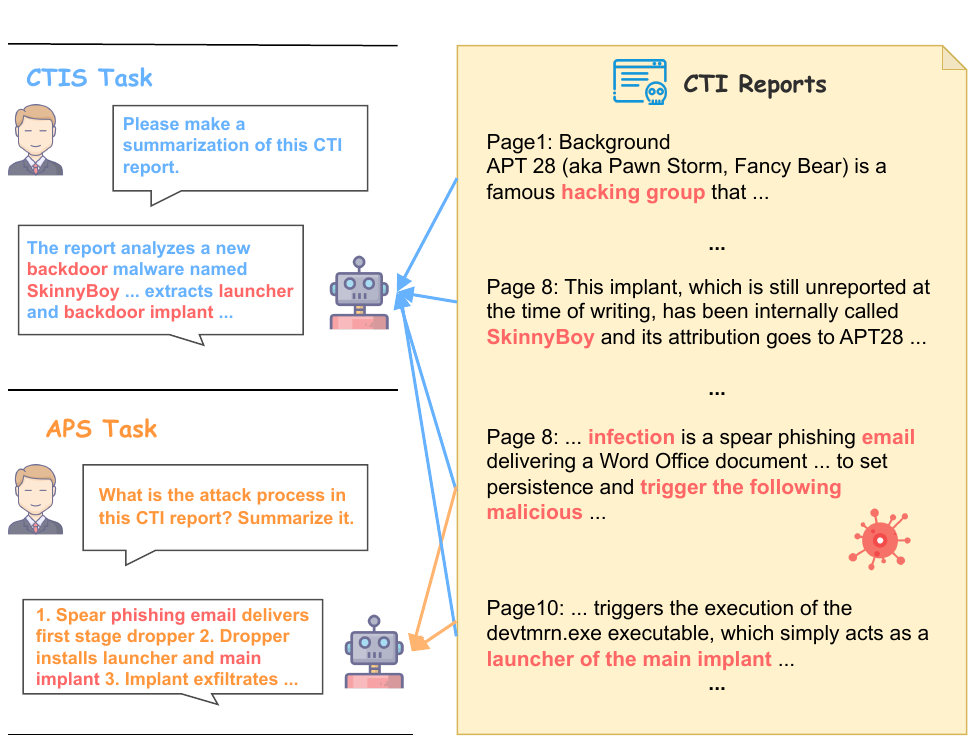}
	\caption{An example in the CTISum. {\color{Blue2} Blue} arrow and {\color{orange} orange} arrow indicate the CTIS and APS tasks, respectively. {\color{red} Red} font represents the key words.
	}
	\label{fig:example}
\end{figure}

With the ongoing advancement of summarization techniques, numerous datasets have emerged, such as CNN/DailyMail \cite{nallapati2016abstractive} and XSum \cite{DBLP:conf/emnlp/NarayanCL18}, among others. In addition, the introduction of these datasets further drives the development of associated technologies like PGNet \cite{DBLP:conf/acl/SeeLM17}, BART \cite{DBLP:conf/acl/LewisLGGMLSZ20}, BRIO \cite{DBLP:conf/acl/LiuLRN22}, etc. 
Furthermore, domain-specific summarization researches have gradually arisen. Examples include biomedicine \cite{DBLP:journals/titb/ZhuYWZ23}, finance \cite{DBLP:conf/emnlp/MukherjeeBBSHSS22}, law \cite{DBLP:conf/ijcnlp/ShuklaBPMGGG22}, etc. This further facilitates various applications involving aiding medical decision-making, generating financial reports, and summarizing legal documents. The boom of domain-specific summarization promotes the understanding of domain terminology and context, which is crucial for generating high-quality summaries for further research. 

Despite remarkable progress made in automatic text summarization, effective methods for summarizing CTI reports in the cybersecurity domain remain largely unexplored. This is primarily due to the lack of available dataset. The unique characteristics of CTI reports, such as technical jargon, evolving threat landscapes, longer reports and fragmented data information, make it challenging to create an annotation, 
which in turn hinders the development of robust models and evaluation benchmarks. 
However, recent advances in Large Language Models (LLMs) \cite{hsiao2023try,cui2024risk,OpenAI2023GPT4TR,touvron2023llama,claudeanthropic,claude2} have shown promising text understanding and generation capabilities. Therefore, this raises the question: \textit{How might we construct a high-quality CTI summarization dataset with the assistance of LLMs?}

In this paper, we construct a new benchmark CTISum 
based on threat intelligence obtained from diverse web sources in the cybersecurity domain. In addition to the CTI Summarization (CTIS) task, considering the importance of attack process, we propose a novel fine-grained subtask, Attack Process Summarization (APS), to enable defenders to quickly understand reported attack behaviors, assess risks and identify security vulnerabilities. Specifically, a multi-stage annotation pipeline is designed to preprocess web CTI reports and obtain annotation data with the assistance of LLMs, which includes the data collection stage, parsing \& cleaning stage, prompt schema stage and intelligence summarization stage. To further keep the high quality of the CTISum, we additionally employ expert reviews for double checking (details can be seen in Section \ref{map}). 
With proper data, automated systems can help synthesize lengthy CTI reports into concise and accurate summaries for different intelligence analysts. 
A sampled report and reference summary in CTISum are shown in Figure \ref{fig:example}. The system not only requires to make a general summary of the full document (task 1), but also needs the capability to focus on and generate the attack process (task 2). After cleaning and human checking, CTISum obtains 1,345 documents and corresponding summaries.

What makes CTISum a challenging dataset can be presented as follows. First, the average document length is about 2,865 words. Current deep learning techniques are incapable of processing documents surpassing 512/1024 tokens in length (after tokenization), such as GPT1\cite{radford2018improving}, T5\cite{raffel2020exploring}, BART\cite{DBLP:conf/acl/LewisLGGMLSZ20} and so on. 
Although LLMs can deal with longer input, however, fine-tuning LLMs is resource consuming. Moreover, LLMs are generally pre-trained on generic datasets, so they often fall short when performing zero-shot tasks in specific domains, such as cybersecurity. Second, the document-to-summary compression ratio on the two tasks are 14.32 (2865.60 / 200.04) and 22.23 (2865.60/118.27), respectively (details in Table \ref{tab:dataset}). The high compression ratio shows competitiveness with the current long document summarization datasets, requiring systems to be extremely precise in capturing only the most relevant facts from lengthy documents in a minimal number of words. 
Finally, CTISum has another subtask, APS, which means the system should be capable of the ability to capture the fine-grained attack process described in CTI reports.

To demonstrate the challenges posed by the proposed CTISum dataset, we conduct comprehensive experiments comparing a range of extractive and abstractive summarization methods. The automatic and human evaluation indicate that existing approaches still have considerable limitations on the CTISum. The extractive methods struggle to identify salient information from lengthy and complex documents. They tend to produce incomplete summaries. Abstractive techniques face challenges in generating coherent and non-redundant summaries, as well as avoiding hallucinations. Both extractive and abstractive methods achieve low scores on automatic metrics like ROUGE. Our findings highlight the need for continued research to develop summarization techniques that can distill critical information in CTISum. The dataset and baselines will be released for further research.

The contributions can be summarized as follows:
\begin{itemize}
	\item To the best of our knowledge, we make the first attempt to build a new benchmark CTISum with the CTIS task and a novel APS subtask in the cybersecurity domain, which focuses on summarizing the key facts or attack process from CTI reports.
	\item A multi-stage annotation pipeline is designed to obtain the high-quality dataset with the assistance of LLMs while manually controlling the quality, which consists of the data collection stage, parsing \& cleaning stage, prompt schema stage and intelligence summarization stage.
	\item Comprehensive experiments on CTISum demonstrate the challenge of the proposed two tasks, meanwhile indicating a large space for future research.
\end{itemize}

\begin{figure*}[!]
	\centering
	\includegraphics[width=0.99\textwidth]{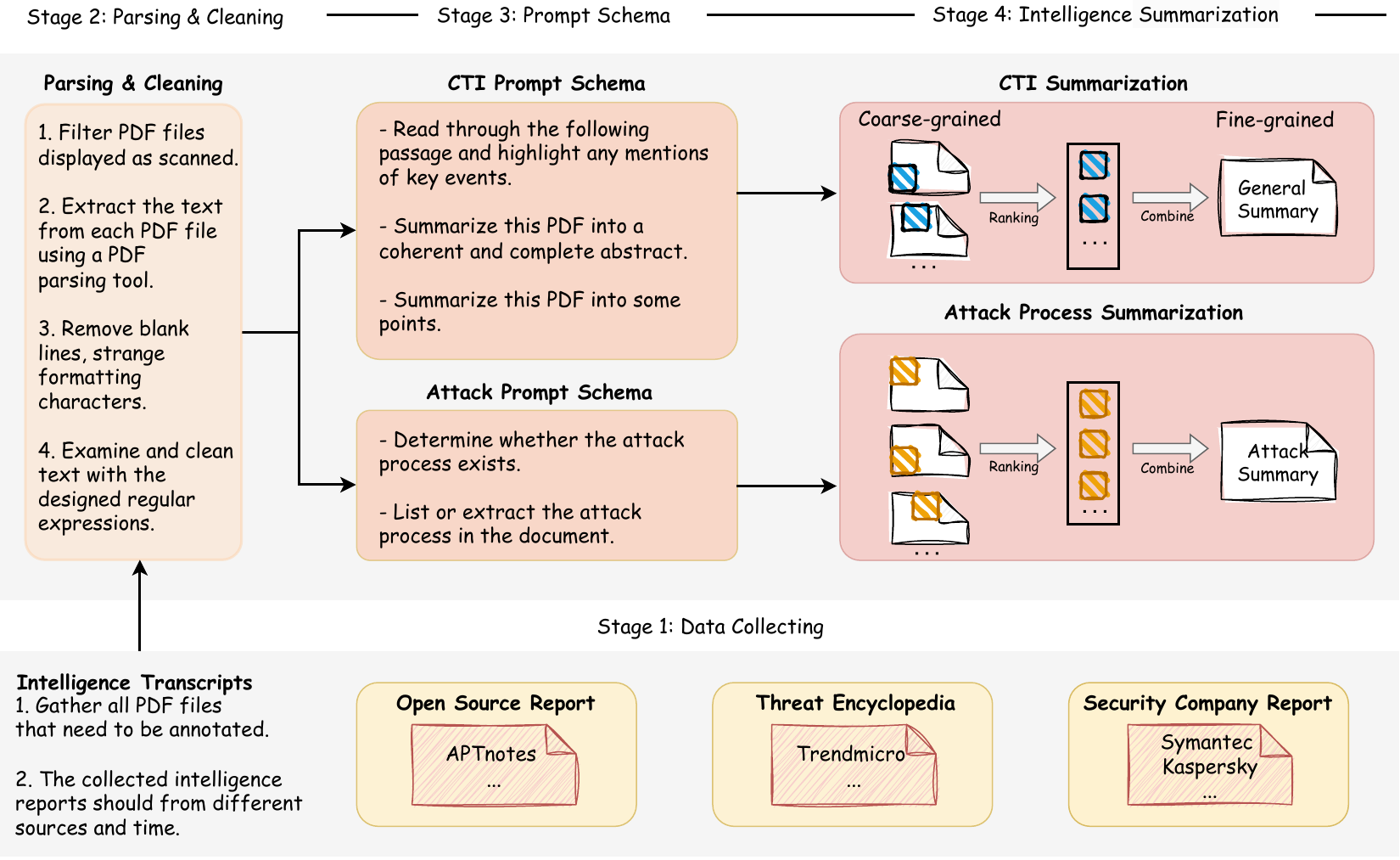}
	\caption{The overview of the proposed multi-stage annotation pipeline for CTIS and APS tasks. }
	\label{fig:model}
\end{figure*}
\section{Related Work}
\subsection{Text Summarization}
A considerable amount of existing researches in automatic text summarization \cite{rush2015neural,nallapati2016abstractive,nallapati2017summarunner,celikyilmaz2018deep,lebanoff2019scoring,DBLP:conf/acl/ZhongLCWQH20} have explored to improve summary generation for news articles, using popular datasets such as CNN/DailyMail \cite{nallapati2016abstractive}, XSum \cite{DBLP:conf/emnlp/NarayanCL18}, etc. In addition, scientific document summarization has emerged as another critical area, with datasets based on ArXiv/PubMed \cite{DBLP:conf/naacl/CohanDKBKCG18}, and other academic sources. The main approaches cover two aspects. First, extractive methods \cite{DBLP:journals/jair/ErkanR04,DBLP:conf/acl/0001L16,DBLP:conf/acl/ZhaoZWYHZ18,DBLP:journals/corr/abs-1903-10318,DBLP:conf/emnlp/LiuL19,DBLP:conf/acl/ZhongLCWQH20} involve identifying and extracting key sentences or passages from the original text to form the summary. For instance, Liu et al. \cite{DBLP:conf/emnlp/LiuL19} introduce a document-level encoder based on BERT and propose a general framework for summarization. Second, abstractive techniques focus on generating new sentences to capture the meaning of the source text. For example, sequence-to-sequence models with attention \cite{DBLP:conf/emnlp/LuongPM15,nallapati2017summarunner,DBLP:journals/kbs/PengHYXX21} are usually utilized to generate the probability distributions on the vocabulary. In addition, See et al. \cite{DBLP:conf/acl/SeeLM17} present a hybrid pointer generator architecture for abstractive summarization. And more recently, transformer-based models like T5 \cite{raffel2020exploring}, BART \cite{DBLP:conf/acl/LewisLGGMLSZ20} and Longformer\cite{beltagy2020longformer} have shown advanced performance on the summarization task. 

Although current general summarization techniques have made some progress, their performance significantly decrease when transferred to CTI summarization tasks due to the lack of domain-specific fine-tuning data. The creation of the CTISum could provide an essential resource for domain adaptation, as well as developing and evaluating domain-specific summarization approaches.

\subsection{Domain-specific Summarization}
For the domain-specific summarization, there are many potential applications across different domains like government \cite{DBLP:conf/naacl/HuangCPJW21}, finance \cite{DBLP:conf/emnlp/MukherjeeBBSHSS22}, court rulings \cite{DBLP:conf/ijcnlp/ShuklaBPMGGG22} and biomedicine \cite{DBLP:journals/titb/ZhuYWZ23}. 
For example, in the government domain, summarization systems can help process large volumes of regulatory text and legal documents. Huang et al. \cite{DBLP:conf/naacl/HuangCPJW21} propose a novel efficient encoder-decoder attention with head-wise positional strides to effectively double the processed input sequence size. For the finance industry \cite{DBLP:conf/emnlp/MukherjeeBBSHSS22}, summarization of earnings reports, financial statements, and news can assist investors and analysts in making decisions. In addition, summarizing court rulings, case law, and litigation documents \cite{DBLP:conf/ijcnlp/ShuklaBPMGGG22} helps legal professionals in their work. Within biomedicine \cite{DBLP:journals/titb/ZhuYWZ23}, healthcare professionals can make informed decisions by using summaries of clinical trial findings, medical literature, and patient records. Different from the above domains, we make the first attempt to build the CTI summarization dataset in the cybersecurity domain, which helps fill the gap and enables new research into CTI summarization techniques.

\section{Data Construction}

As shown in Figure \ref{fig:model}, the proposed multi-stage annotation pipeline consists of four stages. The first stage is collecting data from diverse web sources, namely open source report \cite{aptcyberMonitor}, threat encyclopedia \cite{encyclopedia}, and security company report \cite{symantec} to ensure the diversity and coverage. The second stage is parsing and cleaning which aims to parse original PDF reports into readable text, then obtain high-quality input data by manually cleaning and filtering. Next follows the prompt schema stage, where domain experts design various prompts based on different tasks, to fully utilize the powerful text generation capabilities of LLMs and prepare for the annotation. Finally, the intelligence summarization stage. To combine system productivity and expert experience efficiently, we transform the manual annotation process into ranking and classifying, having experts review and rank the outputs of LLMs to obtain the final summary. Next, we first introduce the multi-stage annotation pipeline, and then make a statistics and analysis of the CTISum.

\subsection{The Multi-stage Annotation Pipeline}
\label{map}
The multi-stage annotation pipeline focuses on obtaining the high-quality CTISum dataset, where we attempt to combine the strengths of both expert experience and LLMs (like GPT-4 \cite{OpenAI2023GPT4TR} or Claude2 or 3.5 \cite{claude2}) to achieve semi-automatic summarization annotation for efficiency in the cybersecurity domain.

\paragraph{Stage 1: Data Collecting.}

To gather relevant CTI reports and build CTISum, we use the reports scraped from diverse web sources, including web open source report like APTnotes\footnote{https://github.com/blackorbird/APT\_REPORT}, threat encyclopedia like Trendmicro\footnote{https://www.trendmicro.com}, and security company report like Symantec\footnote{https://www.broadcom.com/support/security-center} to ensure the diversity and coverage of the CTISum. The range of collected data is covered from 2016 to 2024. For the summarization tasks, a balanced dataset of 1,345 documents is sampled and annotated with two categories: CTIS and APS.

\paragraph{Stage 2: Parsing and Cleaning.}
In order to obtain high-quality input data, the second stage focuses more on parsing and further cleaning the parsed text. 

Given that CTI reports are currently formatted as PDF files, which have complex structure and cannot be directly used as the input to the system, one necessary step is to parse the PDFs and extract readable text. Specifically, the gathered PDFs are firstly filtered to only include those containing actual text content, rather than scanned documents which cannot be parsed by certain tools like PDFMiner. Then, an off-the-shelf PDF parsing library\footnote{https://github.com/euske/pdfminer/tree/master} 
is leveraged to programmatically extract the readable text from each document, while figures are discarded. Finally, the output text is saved in TXT format, allowing transferability to downstream systems.

After parsing, it is indispensable to clean and filter the extracted text for high-quality input data to the LLMs. This post-processing helps improve the quality and consistency of the textual data. One methodology is to manually review samples of the extracted text to identify issues that need to be cleaned. Some rules are:
\begin{compactitem}
	\item Correcting encoding issues when parsing error.
	\item Removing blank lines and odd characters that get extracted but provide no value.
	\item Cleaning up irregular text like ellipses and page numbers that are extracted from tables of contents, lists, etc.
	\item Removing non-English and useless characters.
	\item Deduplicating and removing strings like long sequences of consecutive IP addresses, hashes, etc.
	\item Filtering with regular expressions. Details in Appendix \ref{app:reg}.
\end{compactitem}

The parsing and cleaning stage ensures higher quality input text for the next stage of analysis and annotation.

\paragraph{Stage 3: Prompt Schema.}

The purpose of this stage is to develop a detailed prompt schema to guide the LLMs (Claude2 or 3.5, ChatGLM3, etc.) in producing candidate summaries, such as highlighting key events and summarizing reports, attack processes, some useful annotations, etc.
To generate accurate and consistent summaries for the CTIS task and the APS task, CTI experts analyze the parsed and cleaned text and then design the CTI prompt schema and attack process prompt schema, respectively. As shown in Figure \ref{fig:model}, several key annotation prompts are defined as follows:

\begin{compactitem}
	\item Key Events: The instruction, \textbf{read through the following passage and highlight any mentions of key events}, will focus on important cyber threat events like new malware campaigns, critical vulnerabilities, or notable hacks.
	\item Coherent Summary: To obtain the complete and coherent abstract, the instruction, \textbf{summarize this PDF into a coherent and complete abstract}, is leveraged. 
	\item Additional Points: The purpose of the instruction \textbf{summarize this PDF into some points} is to generate some additional aspects for supplementary.
	\item Determination: Considering some reports do not contain the attack process, the instruction \textbf{determine whether the attack process exists} is designed.
	\item Attack Process: The instruction, \textbf{list or extract the attack process in the document}, is utilized for the APS task.
\end{compactitem}

The prompt schema outlines each annotation prompt in detail, which is a critical component that will guide the LLMs to produce useful, accurate, and consistent summaries.

\paragraph{Stage 4: Intelligence Summarization.}
To combine system productivity and expert experience more efficiently, we transform the manual annotation process into ranking and classifying model-generated results, having three experts review and rank the outputs into overall summaries. Specifically, LLMs like GPT-4o \cite{OpenAI2023GPT4TR}, Claude2 or 3.5 \cite{claude2}, ChatGLM3 \cite{zeng2022glm} are leveraged to generate multiple coarse-grained summaries for the proposed two tasks. Then, three domain experts are involved in reviewing and ranking the candidate summaries, discussing and selecting the better content to combine into a final gold summary. If the generated summaries turn out to be inadequate, the example would be discarded. The domain experts play a key role in evaluating and improving the model-generated summaries to produce a high-quality benchmark. Their human judgment and domain expertise complement the capabilities of LLMs. This collaborative human-AI approach allows the creation of abstractive summaries that are coherent, relevant, and accurate.

\begin{table*}[!thb]
	\centering
	\caption{Statistics of CTISum dataset and existing different domain summarization datasets. Numbers which are not reported are left blank. The numbers for the datasets marked with \textsuperscript{$\ast$} are copied from the paper \cite{DBLP:conf/emnlp/MukherjeeBBSHSS22}, whereas the ones marked with \textsuperscript{$\dagger$} are copied from the work \cite{DBLP:conf/emnlp/NarayanCL18}. Doc.: document, Avg.: average, Len.: length, APS: attack process summarization.} 
	\resizebox{\linewidth}{!}{
		\begin{tabular}{lcccccc}
			\toprule
			\textbf{Dataset} & \textbf{\# Doc.} & \textbf{Avg. Doc. Len.} & \textbf{Avg. Sum. Len.} & \textbf{Avg. APS Len.} & \textbf{Has Subtask} & \textbf{Domain}\\
			\midrule
			General Domain & & & & & & \\ 
			\textsc{Arxiv/PubMed} \cite{DBLP:conf/naacl/CohanDKBKCG18}\textsuperscript{$\ast$} & 346,187 & 5,179.22 & 257.44 & - & No & Academic  \\
			\textsc{CNN} \cite{nallapati2016abstractive}\textsuperscript{$\dagger$} &  92,579 & 760.50 & 45.70 & - & No & News \\
			\textsc{DailyMail} \cite{nallapati2016abstractive}\textsuperscript{$\dagger$} & 219,506 & 653.33 & 54.65 & - & No & News \\
			\textsc{XSum} \cite{DBLP:conf/emnlp/NarayanCL18}\textsuperscript{$\dagger$} & 226,711 & 431.07 & 23.26 & - & No & News \\
			\textsc{BookSum} Chapters \cite{DBLP:conf/emnlp/MukherjeeBBSHSS22}\textsuperscript{$\ast$} & 12,630 & 5,101.88 & 505.42 & - & No & Books \\
			\midrule
			Specific Domain & & & & & & \\ 
			\textsc{GovReport} \cite{DBLP:conf/naacl/HuangCPJW21} & 19,466 & 9,409.40 & 553.40 & - & No & Government  \\
			\textsc{Discharge} \cite{DBLP:journals/titb/ZhuYWZ23}\ & 50,000 & 2,162.29 & 28.84 & - & No & Biomedicine  \\
			\textsc{Echo} \cite{DBLP:journals/titb/ZhuYWZ23} & 162,000 & 315.30 & 49.99 & - & No & Biomedicine \\
			\textsc{ECTSum} \cite{DBLP:conf/emnlp/MukherjeeBBSHSS22} & 2,425 & 2,916.44 & 49.23 & - & No & Finance  \\
			\textsc{IN-Ext} \cite{DBLP:conf/ijcnlp/ShuklaBPMGGG22} & 50 & 5,389 & 1,670 & - & No & Court Rulings \\
			\textsc{UK-Abs} \cite{DBLP:conf/ijcnlp/ShuklaBPMGGG22} & 793 & 14,296 & 1,573 & - & No & Court Rulings\\
			\midrule
			\textsc{CTISum} & 1,345 & 2,865.60 & 200.04 & 118.27 & Yes & Cybersecurity \\
			\bottomrule
	\end{tabular}}
	\label{tab:dataset}
\end{table*}

\subsection{Data Statistics}
Table \ref{tab:dataset} describes the statistics of the CTISum dataset and existing summarization datasets from different domains. The datasets are divided into general and specific domain categories, with the latter attracting substantial interest in recent years from 2021 to 2024. And we make the first attempt to build the CTI summarization dataset in the cybersecurity domain. Specifically, CTISum has 1,345 documents from diverse web sources, with an average document length of 2,865 words, with an average CTI summary length of 200 words, and with a high compression ratio of 14.32. Notably, CTISum is the only dataset with the subtask, while other datasets simply make a general summary of the document. The average length of the attack process summaries is about 118 words, which is shorter than the general summaries. This highlights the challenging nature of generating focused summaries that capturing the attack process from long documents. In addition, the Fleiss Kappa evaluation of the dataset can be seen in Sec. \ref{app:kappa}.

In summary, the table highlights that CTISum is a challenging dataset for summarization in the cybersecurity domain, with a novel subtask (attack process summarization) and high abstraction requirements. The CTISum dataset helps fill the gap and enables new system design in future work.

\section{Problem Formulation}
Existing approaches usually define the summarization task as a sequence-to-sequence task. Specifically, the problem formulation can be formulated as follows. Given a document $D = (x_{1}, \dots, x_{N})$ that consists of ${N}$ words, with $Y = (y_1, \dots, y_M)$ being the corresponding summary. For the proposed CTISum, there are two tasks, including CTIS task and APS task. For CTIS task, the object is to output a general summary of the CTI report. Similarly, the object of the APS task is to generate a summary of the attack process.

\section{Experiments}
\subsection{Evaluation Metrics}
In this section, the automatic and human A/B evaluation are considered to validate the performance of the current SOTA models.

Following papers \cite{DBLP:conf/emnlp/HsuS022,DBLP:conf/emnlp/MukherjeeBBSHSS22}, BERTScore \cite{DBLP:conf/iclr/ZhangKWWA20} and ROUGE-$n$ (R-$n$) \cite{lin2004rouge} are taken as evaluation metrics, which are widely used for evaluating the quality of summarization. BERTScore is a metric for evaluating text generation models, which measures the similarity between the generated text and reference text using contextual embeddings from pre-trained BERT. 
ROUGE-$n$ refers to the overlap of n-grams between the generated and reference summaries. Specifically, ROUGE-$1$ evaluates unigram overlap, ROUGE-$2$ measures bigram overlap, and ROUGE-$L$ calculates longest common subsequence overlap.
These metrics compare matching units such as words and phrases between the generated and reference summaries, with higher scores indicating better summarization quality. 
Among the ROUGE metrics, ROUGE-$L$ generally corresponds best with human judgments of summary quality. 

In previous studies, human evaluation is usually conducted by crowdsourcing workers who rate responses on a scale from 1 to 5 from the aspects of correctness, relevancy, etc. However, {the} criteria can vary widely between different individuals. Therefore, this study adopts the human A/B evaluation for a high inter-annotator agreement. Given the generated summaries of two models A and B, three analysts are prompted to go through an entire CTI report and choose the better one for each of the 80 randomly sub-sampled test instances.
For objectivity, annotators include those with and without background knowledge (task-related). 
The final results are determined by majority voting. 
If the three annotators reach different conclusions, the fourth annotator will be brought in.
We adopt the same human evaluation with paper \cite{DBLP:conf/emnlp/MukherjeeBBSHSS22}: 1) Factual Correctness: which summary can be supported by the source CTI report? 2) Relevance: which summary captures pertinent information relative to the CTI report? 3) Coverage: which summary contains the greatest coverage of relevant content in the CTI report?


\begin{table*}[t]
	\centering
	\setlength\tabcolsep{6pt}
	\caption{\label{tab:1} Performance of automatic evaluation on CTIS and APS tasks. LLAMA2 is in the zero-shot setting because of the lack of resources. The best results are highlighted in \textbf{bold}.}
	\resizebox{\linewidth}{!}{
		\begin{tabular}{lcccccccc}
			\toprule
			& \multicolumn{4}{c}{\textbf{CTIS  Validation}}
			& \multicolumn{4}{c}{\textbf{CTIS Test}}  \\
			\textbf{Model}   & \textbf{BERTScore}$\uparrow$		& \textbf{R-1}$\uparrow$	& \textbf{R-2}$\uparrow$	& \textbf{R-L}$\uparrow$	& \textbf{BERTScore}$\uparrow$		& \textbf{R-1}$\uparrow$ & \textbf{R-2}$\uparrow$ & \textbf{R-L}$\uparrow$ \\
			\midrule	
			\textbf{Extractive} \\
			BertSumExt \cite{DBLP:conf/emnlp/LiuL19}   & 82.15  &   26.91 &  5.91 & 13.09 & 81.36  &  20.43 &  3.52 & 11.76            \\
			MatchSum \cite{DBLP:conf/acl/ZhongLCWQH20}    & \textbf{83.91}  &   39.31 & 13.39 & 20.60 & \textbf{83.75}  &  33.76 &  9.34 & 18.91             \\
			\midrule
			\textbf{Abstractive} \\
			Transformer \cite{DBLP:conf/nips/VaswaniSPUJGKP17} & 45.83  &   34.15 &  12.05 & 19.88 & 45.17  &   32.94 &  10.37 & 18.55           \\
			T5 \cite{raffel2020exploring}   & 70.48 & 43.75  & 16.58 & 28.06 & 69.21 & 42.41 & 14.35 & 27.32 \\
			BART-base \cite{DBLP:conf/acl/LewisLGGMLSZ20}   & 70.36 & 44.45  & 16.26 & 27.24 & 69.24 & 42.48 & 14.30 & 26.23 \\
			\textbf{BART-large} \cite{DBLP:conf/acl/LewisLGGMLSZ20}   & {71.21} & \textbf{47.11} & \textbf{18.32}  & \textbf{29.20} & {70.41} & \textbf{45.76} & \textbf{16.88} & \textbf{29.03} \\
			\midrule
			\textbf{Long-document-based} \\
			Longformer\cite{beltagy2020longformer}  	& 71.06 & 46.97 & 17.21 & 28.22 	& 70.31 & 45.39 & 15.66 & 27.09    \\
			\textbf{LLAMA2} \cite{touvron2023llama} (zero-shot)	   & 33.87  &   21.35 &  5.93 & 20.51 &  32.15 &   20.89 &  5.66 & 19.22        \\
			\textbf{GPT-4o} (zero-shot)	   &  65.82 & 40.18  &   13.22  & 24.33 &  64.31 &   38.70 &  12.71 & 22.98   \\
			\toprule
			\midrule
			& \multicolumn{4}{c}{\textbf{APS  Validation}}
			& \multicolumn{4}{c}{\textbf{APS Test}}  \\
			\textbf{Model}   & \textbf{BERTScore}$\uparrow$		& \textbf{R-1}$\uparrow$	& \textbf{R-2}$\uparrow$	& \textbf{R-L}$\uparrow$	& \textbf{BERTScore}$\uparrow$	& \textbf{R-1}$\uparrow$ & \textbf{R-2}$\uparrow$	& \textbf{R-L}$\uparrow$\\
			\midrule	
			\textbf{Extractive} \\
			BertSumExt \cite{DBLP:conf/emnlp/LiuL19}    & 81.38  &   20.67 &  3.06 & 11.87 & 81.38  &   20.35 &  3.46 & 11.73      \\
			MatchSum \cite{DBLP:conf/acl/ZhongLCWQH20}    & \textbf{83.93}  &   33.86 &  9.86 & 19.17 & \textbf{83.75}  &   33.73 &  9.35 & 18.88    \\
			\midrule
			\textbf{Abstractive} \\
			Transformer \cite{DBLP:conf/nips/VaswaniSPUJGKP17}  & 40.43  &   28.22 &  7.86 & 16.50 & 39.15  &   27.03 &  6.21 & 15.48   \\
			T5 \cite{raffel2020exploring}   & 68.41 & 36.20 & 10.27 & 24.70 & 66.06 & 32.09 & 7.62  & 21.78 \\
			BART-base \cite{DBLP:conf/acl/LewisLGGMLSZ20}   &  70.47 & 40.47 & 12.33 & 25.28 & 70.70 & 39.42 & 11.13 & 24.45  \\
			\textbf{BART-large} \cite{DBLP:conf/acl/LewisLGGMLSZ20}   & {71.31} & \textbf{41.88} & \textbf{12.76} & \textbf{26.54} & {70.32} & \textbf{40.25} & \textbf{11.65} & \textbf{25.06}   \\
			\midrule
			\textbf{Long-document-based} \\
			Longformer\cite{beltagy2020longformer}  		& 70.98 & 41.35 & 11.77 & 25.36 & 70.41 & 39.91 & 9.78  & 23.93 \\
			\textbf{LLAMA2} \cite{touvron2023llama} (zero-shot)	   & 21.06  &   15.68 &  2.23 & 13.14 &  20.38 &   14.66 &  2.02 & 12.28   \\
			\textbf{GPT-4o} (zero-shot)	   & 48.77  &   32.06 &  9.11 & 18.69 & 45.39  &   30.55 &  8.61 & 17.75   \\
			\bottomrule
	\end{tabular}}
\end{table*}

\subsection{Experimental Setting}
The implementation of baselines is based on the HuggingFace framework. The AdamW optimizer \cite{Loshchilov2017FixingWD} with $ \beta_1 = 0.9$ and $\beta_2 = 0.99$ is used for training, with an initial learning rate of $3e-5$ and a linear warmup with $100$ steps. The batch size is set to $16$ for training, and we use a batch size of $1$ and a maximum of $128$ decoding steps during inference. 
Top-$p$ sampling is set to $0.9$, temperature $\tau=0.7$. The epoch is set to $5$. For preprocessing, we randomly split the dataset into train, validation and test set with a ratio of 8:1:1.

\subsection{Baselines}
Several SOTA approaches are illustrated for comparison. The models can be mainly divided into extraction-based, abstraction-based and long-document-based methods.

\paragraph{Extractive Model}
Extractive summarization involves selecting a subset of salient sentences, phrases, or words from the original document to form the summary. The key idea is to identify and extract the most important content, and then arrange extracted content to flow logically. We simply introduce some methods as follows.

BertSumExt \cite{DBLP:conf/emnlp/LiuL19} is a neural extractive summarization model based on BERT \cite{DBLP:conf/naacl/DevlinCLT19}, which takes BERT as the sentence encoder and a Transformer layer as the document encoder. A classifier is utilized to perform sentence selection, then the model outputs the final summary. 

MatchSum \cite{DBLP:conf/acl/ZhongLCWQH20} formulates the extractive summarization task as a semantic text matching problem and develops a novel summary-level framework MatchSum, which generates a collection of possible candidate summaries from the output of BertSumEXT. The candidate that matches best with the document is selected as the final summarized version.

\paragraph{Abstractive Model}
Abstractive summarization involves generating new phrases and sentences that convey the most important information from the source document, which uses semantic understanding and language generation technology to make a summarization. Hence, we fine-tune BART \cite{DBLP:conf/acl/LewisLGGMLSZ20} and T5 \cite{raffel2020exploring} from the HuggingFace library.

BART \cite{DBLP:conf/acl/LewisLGGMLSZ20} is an encoder-decoder language model based on a sequence-to-sequence architecture, which achieves strong performance on a variety of NLP tasks 
like summarization, question answering and text generation after fine-tuning. 

T5 \cite{raffel2020exploring} uses a standard Transformer encoder-decoder architecture, which converts all NLP tasks into a unified text-to-text format where the input and output are always text strings. T5 comes in several sizes including T5-Small, T5-Base, T5-Large, etc. In this paper, the setting is based on T5-Base.

\paragraph{Long-document-based Model}
Long-document-based summarization refers to generating summaries for documents that are significantly longer than typical texts, e.g. 1,024 tokens, like Longformer\cite{beltagy2020longformer}, LLAMA2 \cite{touvron2023llama}, etc.

Longformer\cite{beltagy2020longformer} is an extension to the Transformer architecture, which leverages an attention mechanism whose computational complexity grows linearly as the length of the input sequence increases, making it easy to process thousands of tokens or more. LLAMA2 \cite{touvron2023llama} is a Transformer-based large language model, which is trained on massive text datasets to learn natural language patterns and generate human-like text.
The context length of LLAMA2 can extend up to 4096 tokens, allowing it to understand and generate longer texts. In this paper, the 7 billion LLAMA2 is utilized to compare against fine-tuned models in zero-shot setting.

\begin{table*}[h]
	\setlength\tabcolsep{9pt}
	\centering
	\caption{Results for the human evaluation of model-generated summaries by three intelligent analysts.}
	\resizebox{\linewidth}{!}{
		\begin{tabular}{lccccccc}
			\toprule
			& \multicolumn{3}{c}{\textbf{CTIS  Task}}
			& \multicolumn{4}{c}{\textbf{APS Task}}  \\
			& \textbf{Correctness} & \textbf{Relevance} & \textbf{Coverage} & & \textbf{Correctness} & \textbf{Relevance} & \textbf{Coverage} \\
			
			\toprule
			
			\textbf{Model} & \multicolumn{7}{c}{\textbf{Summary-level scores (about 80 summaries)}} \\
			
			\midrule
			
			\textbf{BART better} & 33.33  & 26.67  & 46.67   &		& 40.00  & 33.33  & 46.67	\\
			\textbf{Longformer better} & 26.67  & 20.00  & 20.00  &	& 26.67  & 26.67  & 20.00 \\
			\textbf{Both equally good} & 40.00  & 53.33  & 33.33  &	& 33.33  & 40.00  & 33.33 \\
			
			\bottomrule
		\end{tabular}
	}
	\label{tab:human_eval}
\end{table*}

\subsection{Main Results}
\paragraph{Automatic Evaluation.}
We compare the performance of extractive models, abstractive models and long-document-based models on CTIS and APS tasks. As depicted in Table \ref{tab:1}, extractive models perform the almost worst on both tasks, demonstrating that intelligence summarization in cybersecurity involves more than just extracting sentences. Interestingly, they obtain the best BERTScore, one possible reason is that these two models are based on BERT, leading to high BERTScore. Compared with MatchSum, abstractive model like BART-large (about 374M) achieves better result, about 10.12\% and 6.18\% gain on ROUGE-L on two tasks, which shows that the abstractive models are more suitable for these two tasks. Then, long-document-based model like Longformer (102M) obtains similar results compared with BART-base (about 121M). 
Notably, LLAMA2-7B and GPT-4o (zero-shot) do not attain further improvement on all the evaluation metrics, leaving noticeable room for future work on domain LLMs fine-tuning and few-shot learning on the benchmark. Finally, we find that performance is worse on the APS task, likely because the attack process exhibits properties like variability, concealment and complexity, making APS more challenging than general summarization. Enhancing models for APS task remains an important direction for future research.

\begin{figure*}[t]
	\centering
	\begin{tabular}{cc}
		\includegraphics[width=0.49\textwidth, height=0.28\textwidth]{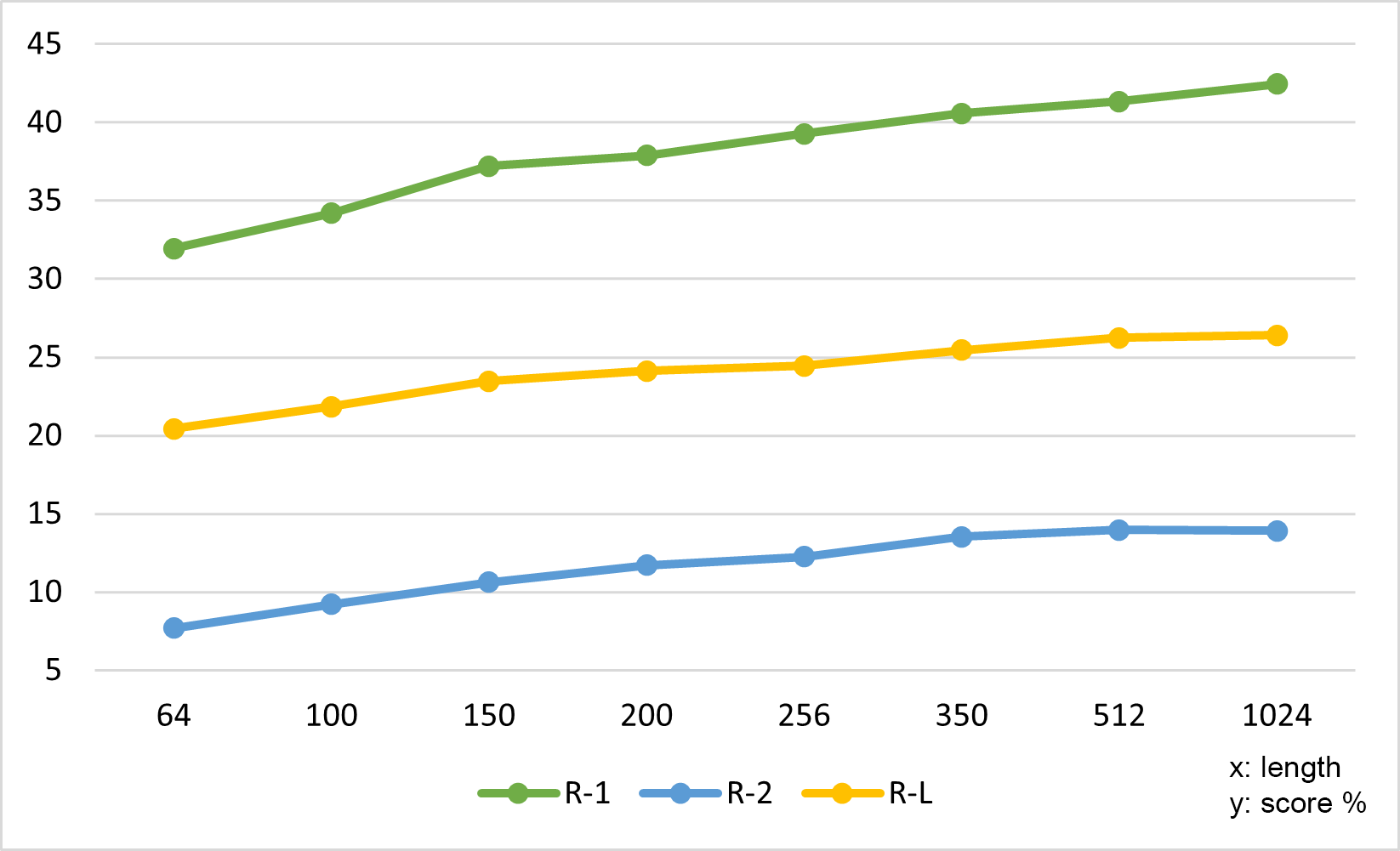} &
		\includegraphics[width=0.49\textwidth, height=0.28\textwidth]{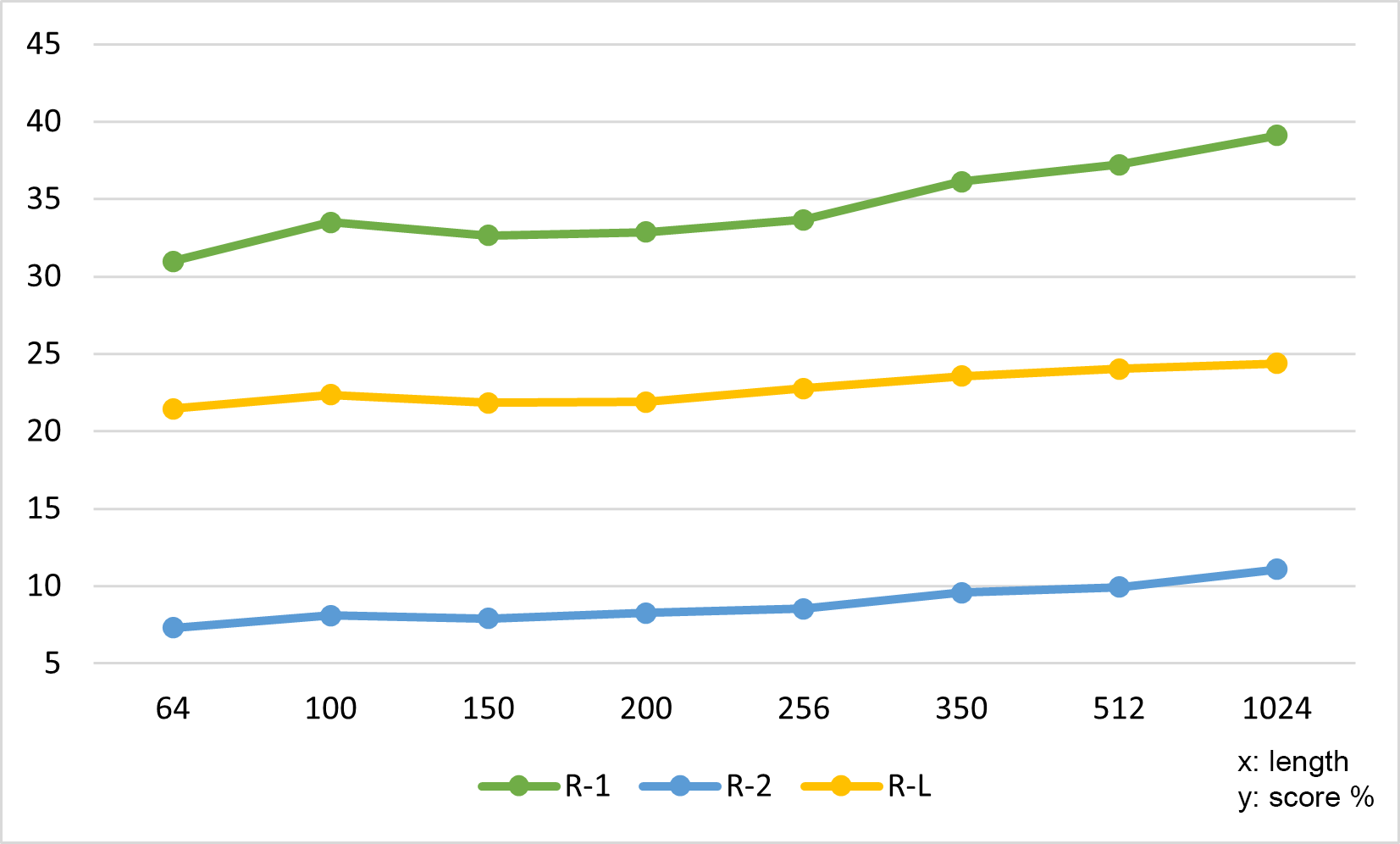} \\
		\textbf{(a) CTIS task.} & \textbf{(b) APS task.} \\
	\end{tabular}
	\caption{Input length analysis of two tasks.}
	\label{fig:led_vs_our}
\end{figure*}

\begin{table}[t]
	\centering
	\caption{Fleiss Kappa evaluation of CTISum dataset.} 
	\resizebox{0.95\linewidth}{!}{
		\begin{tabular}{l|ccccc|c}
			\toprule
			\textbf{Score} & \textbf{1} & \textbf{2} & \textbf{3} & \textbf{4} & \textbf{5} & \textbf{Fleiss Kappa}\\
			\midrule
			Result & 0 & 9 & 58 & 129 & 54 & 0.37 \\ 
			\bottomrule
			\textbf{Score}  & \multicolumn{2}{c}{\textbf{1 and 2}} & \textbf{3}  & \multicolumn{2}{c|}{\textbf{4 and 5}} & \textbf{Fleiss Kappa} \\
			\midrule
			Result & \multicolumn{2}{c}{9}       & 58 & \multicolumn{2}{c|}{183}     & 0.61    \\
			\bottomrule
	\end{tabular}}
	\label{tab:kappa}
\end{table}

\paragraph{Human A/B Evaluation.}
In addition to the automatic evaluation, human A/B evaluation \cite{DBLP:journals/kbs/PengQHXL23} is conducted to validate the effectiveness of current SOTA models that BART-large and Longformer are compared. The results in Table \ref{tab:human_eval} demonstrate the consistent conclusion with automatic evaluation. 
It can be seen that the summaries from BART-large are much more preferred than those of the baselines in terms of the three aspects. For example, compared with Longformer, BART is superior in terms of the coverage metric, indicating that the larger the number of parameters, the better the generated summaries. Besides, it is noted that BART does not significantly outperform Longformer in the relevance metric, and this is probably attributed to the powerful understanding ability of the {PLMs}, but BART still obtains decent improvements. 

\subsection{Fleiss Kappa Analysis}
\label{app:kappa}
To verify the consistency of the final summary generation quality, we select 5 experts to execute a consistency evaluation of 50 sampled summaries using Fleiss Kappa, where the scores are rated from 1 to 5, with 1 very negative, 2 negative, 3 neutral, 4 positive, and 5 very positive. As shown in the Table \ref{tab:kappa}, the final Fleiss Kappa score is 0.37, indicating relatively low agreement. This can be attributed to the fact that summary generation belongs to language generation task, where different experts tend to have subjective views on summary quality (e.g., examples that annotated as positive or very positive are inconsistent, but they are all good), which differs greatly from the objectivity of classification tasks. However, we also analyze the score distribution of all experts, finding that summaries rated 3 or above account for 96.4\% (241/250) of the total, demonstrating that the quality of the annotated dataset is controllable and accurate. Moreover, 
we have conduct another round of Fleiss Kappa Analysis on the previous annotation scores from multiple raters. Specifically, we combine score 1 and 2 into one bin, keep score 3 as a separate bin, and combine score 4 and 5 into another bin. This result in a Fleiss Kappa score of 0.613184, which indicate the consistency evaluation.
\begin{table*}[t]
	\setlength\tabcolsep{8pt}
	\caption{Case analysis on CTIS and APS tasks. \textbf{\color{orange}{Orange words}} mean the hallucination generated by the baseline.  \textbf{\color{Blue2}{Blue words}} present the correct result. \textbf{\color{red}{Red words}} indicate some of the key information that needs to be extracted.} 
	\begin{tabular}{p{1.8cm}|p{14.35cm}}
		\toprule
		{Task}   & \begin{tabular}{c} CTIS Task  \end{tabular} \\
		\midrule
		{Document}   & \begin{tabular}[c]{@{}l@{}} ShadowGate Returns to Worldwide Operations With Evolved \textbf{\color{red}Greenﬂash Sundown} ... After almost two \\ years of sporadic restricted activity, the \textbf{\color{red}ShadowGate campaign} has started delivering cryptocurrency \\ miners with a    newlyupgraded version of the Greenﬂash Sundown exploit kit ... \textbf{\color{red}ShadowGate was} \\ \textbf{\color{red}active since 2015} but restricted operations after a takedown \textbf{\color{red}in 2016} ... \textbf{\color{red}In April 2018}, ShadowGate \\ was spotted spreading  ...  
			
		\end{tabular} \\ \midrule
		{BART}   & \begin{tabular}[c]{@{}l@{}} The PDF describes the activities of a threat actor group called ShadowGate ... \textbf{\color{orange}The campaign has been} \\ \textbf{\color{orange} actively}  \textbf{\color{orange}evolving its exploit kit since 2016} ... ShadowGate started distributing cryptocurrency miners  \\ using a new version of \textbf{\color{orange}Green\_Sundown}, likely to avoid ... The Monero  \\
		\end{tabular} \\ \midrule
		{Ground truth}   & \begin{tabular}[c]{@{}l@{}} The report analyzes the resurgence of a cybercrime campaign called ShadowGate, which is using an \\ upgraded version of its custom exploit kit, Greenflash Sundown, to infect victims worldwide with  \\ cryptocurrency mining malware. ShadowGate was active \textbf{\color{Blue2}since 2015} but restricted operations after \\  a takedown \textbf{\color{Blue2}in 2016}.It recently  resumed attacks using compromised ...  \\
			
		\end{tabular}                                                                             \\    
		\bottomrule
		\bottomrule
		{Task}   & \begin{tabular}{c} APS Task  \end{tabular} \\
		\midrule
		{Document}   & \begin{tabular}[c]{@{}l@{}} Staring at a major financial resource crunch, MSME's are worst affected due to the ongoing COVID-19 \\ pandemic. We observed one similar wave on MSME in late April 2020 — it was a phishing campaign luring \\ victims with \textbf{\color{red}COVID-19 themed maldocs} ... Victims \textbf{\color{red}received an email} with attached zip``face mask \\ order.zip'' which contained the \textbf{\color{red}aforementioned maldoc} ... The attack begins \textbf{\color{red} after opening ``face mask}  \\ \textbf{\color{red} order.doc''}. This RTF is weaponized with exploit which triggers \textbf{\color{red}CVE-2017-11882} vulnerability to  \\  execute arbitrary code ... injected Agent Tesla payload in \textbf{\color{red}RegAsm.exe} ...   \\
			
		\end{tabular} \\ \midrule
		{BART}   & \begin{tabular}[c]{@{}l@{}} 1. The phishing email contains a malicious RTF file titled ``face mask order.zip" 2. The RTF contains an \\ exploit that triggers \textbf{\color{Blue2}CVE-2017-11882} to execute arbitrary code. 3. The exploit extracts the \textbf{\color{orange}OLE} objects \\ from RTF files. 4. The OLE object contains a VBScript script that executes an \textbf{\color{orange}Equation Editor} payload.  \\ 5. The Equated Editor payload connects to a command and control server to receive commands. \\  6. Command server sends commands to the C2 server to execute additional payloads. 7. Additional \\ payloads are delivered to the infected ...  \\
		\end{tabular} \\ \midrule
		{Ground truth}   & \begin{tabular}[c]{@{}l@{}} 1. Gorgon APT sends phishing emails with COVID-19 themed Word documents to targets in the MSME \\ sector. 2. The documents contain an exploit for \textbf{\color{Blue2}CVE-2017-11882} to execute malicious code. 3. The code  \\ drops a visual basic script named \textbf{\color{Blue2}ServerCrypted.vbs}. 4. The VBScript executes a \textbf{\color{Blue2}PowerShell command} \\ to download additional payloads. 5. The first payload is an injector DLL. 6. The second payload is the \\  Agent Tesla Remote Access Trojan. 7. The injector DLL loads itself into memory using PowerShell. \\ 8. Agent Tesla is injected into the \textbf{\color{Blue2}RegAsm.exe} process. 9. Agent Tesla ...   \\
			
		\end{tabular}                              \\ 
		\bottomrule
		\noalign{\vskip -1mm}
	\end{tabular}
	\label{compare}
\end{table*}

\subsection{Input Length Analysis}
\label{sec:len}
In this section, the input length analysis is performed to study the model's sensitivity to input length. We evaluate the impact of different input lengths on the BART-base (the purpose of using base model is to train the model quickly) on CTI and APS tasks. We divide the different length to get better insight into the current models. As shown in Figure \ref{fig:led_vs_our}, where the x-axis represents the input length and the y-axis denotes the measure value of evaluation metrics. The conclusions can be drawn: 1) As the input length increases, the model performance improves steadily, highlighting the model's ability to process and summarize longer sequences. 2) As the length of BART reaches 1024 tokens, performance on our tasks continues to improve. To further investigate the impact of longer length, we increase the size of Longformer and find that lengthening to 3500 tokens (75th percentile of length distribution) results in optimal performance, which suggests the advantage of long-distance modeling. 3) The model achieves poorer performance on the APS task compared to the CTIS task, indicating it struggles more with summarizing technical attack details than threat intelligence reports. 
In summary, the experiments provide insight into current models' ability to capture long-distance dependencies in documents and summarize essential information as the input size varies. Further work could focus on improving the model's ability of length to boost performance on lengthy, technical summarization tasks.

\begin{figure*}[t]
	\centering
	\begin{tabular}{cc}
		\includegraphics[width=0.48\textwidth, height=0.27\textwidth]{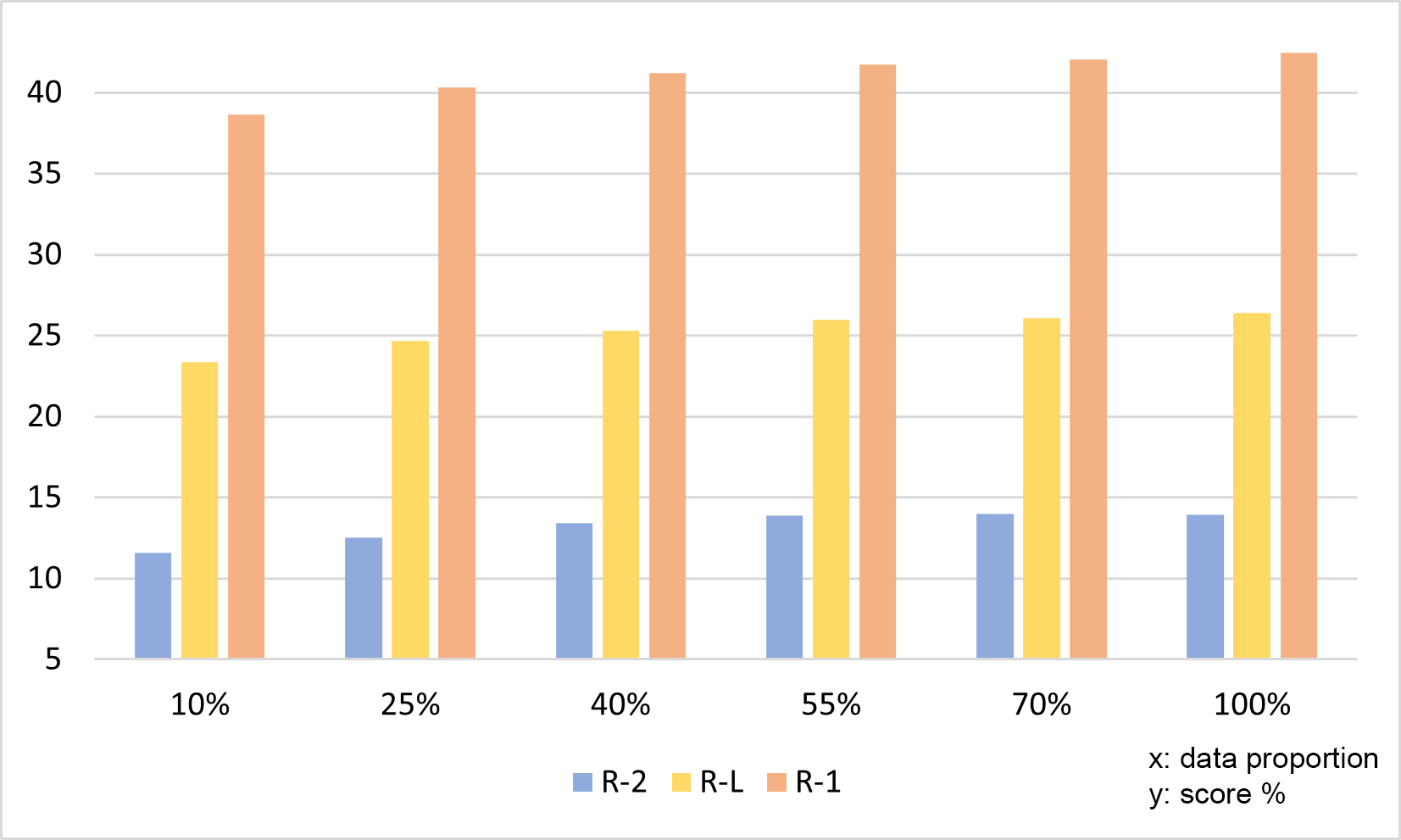} &
		\includegraphics[width=0.48\textwidth, height=0.27\textwidth]{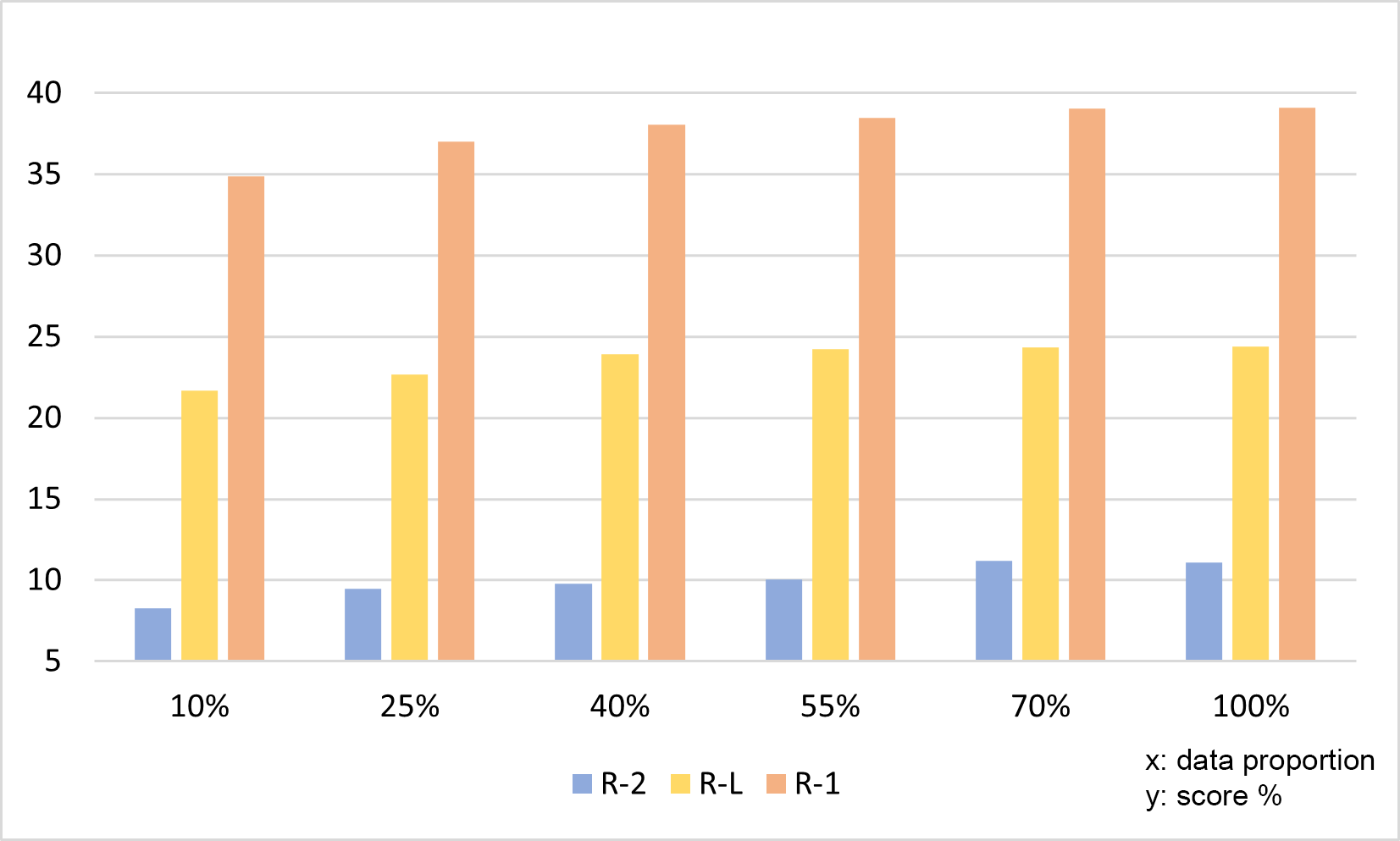} \\
		\textbf{(a) CTIS task.} & \textbf{(b) APS task.} \\
	\end{tabular}
	\caption{Few shot experiments of the proposed two tasks.}
	\label{fig:datasize}
\end{figure*}

\subsection{Few-shot Analysis}
Analyzing the impact of data size can help to understand the relationship between data size and summarizing performance. To demonstrate such a relationship, in this section, few-shot analysis experiments on BART-base are conducted to study model's sensitivity in different training data sizes. We experiment with 10\%, 25\%, 40\%, 55\%, 70\%, and 100\% of the total training data to analyze the impact on CTIS task and APS task. The experimental results are shown in the Figure \ref{fig:datasize} with several conclusions: 1) Performance of BART-base improves as the training data size increases, indicating that model quality benefits from more abundant training data. Furthermore, the result in the figure is still growing, which suggests further improvements with larger training data size. 2) Performance on the APS task is worse than the CTIS task, especially in few shot setting (e.g., 10\%, 25\%). 
Therefore, how to improve the performance of summarization models in few-shot setting for cybersecurity domain will be an important future research track. And data augmentation is also another area worth researching.

\subsection{Case Analysis}
Qualitatively, we observe interesting examples on the BART-base from CTIS and APS tasks in Table \ref{compare}. In case one on CTIS task, the model generates hallucination about time \textit{\color{orange}since 2016}, leading to incorrect time information in the summary. This phenomenon likely occurs because if the source document contains multiple time details, the model may confuse about which time points are being referenced. Therefore, addressing this challenge will be critical for future improvements. In case two on APS task, while generated attack process summary can include some factual information \textit{\color{Blue2}CVE-2017-11882}, many factual inconsistencies errors exist, such as \textit{\color{orange}OLE} and \textit{\color{orange}Equation Editor}. This stems from attack behaviors that exhibit properties of complexity, diversity, and contextual dependence, which makes the APS task difficult. Thus, adapting appropriate methods to better capture the intricacies of attack behaviors represents a promising research direction. 
Overall, these cases highlight important limitations, but also provide exciting opportunities to advance CTIS and APS tasks through innovations targeting core summarize-ability challenges in the cybersecurity domain.

\section{Conclusion}
With a lack of publicly available intelligence data, in this paper, we make the first attempt to propose CTISum benchmark which provides a valuable resource to spur innovation in this critical but unobserved cybersecurity domain. The creation of this intelligence-focused summarization benchmark represents an important step toward developing AI systems that can effectively process and synthesize CTI reports. To combine system productivity and expert experience more efficiently, a multi-stage annotation pipeline is designed for obtaining the high-quality dataset.
Our experiments reveal that while current models can produce decent summaries, there is significant room for improvement in capturing key details accurately. The technical cybersecurity terminology and complex contextual concepts present in the reports pose great challenges for generating intelligence summaries. In the future work, leveraging transfer learning and LLMs to the cybersecurity domain can provide a starting point, as well as exploring adaptations of summarization techniques to address the unique characteristics of the cyber threat intelligence field. Ethical statement is in Appendix \ref{app:eth}.

\section*{Acknowledgments}
We thank all anonymous reviewers for their constructive comments. This work is supported by the Zhongguancun Laboratory.

\appendix
\section{Appendix}

\subsection{Ethical Statement}
\label{app:eth}
All data in CTISum dataset has been double checked by experts that major in intelligence analysis, and manually checked to ensure the absence of any ethical or political inaccuracies.

\subsection{Details of Regular Expression}
\label{app:reg}
During the process of filtering and cleaning the parsed PDF data, in order to reduce the burden of manual checking by annotators, we first sample and observe 50 examples, and then extract some common features to formulate into regular expressions for automatically cleaning the data. Specifically, details of regular expression are in the following:
\begin{itemize}
	\item Removing consecutive dots (table of contents)
	\item Deleting redundant newlines/carriage returns 
	\item Removing extra whitespace between characters
	\item Deleting multiple consecutive lines of IP addresses, hashes, etc.
	\item Removing website links starting with ``http''
	\item Removing Thai, Korean and other non-English characters
\end{itemize}
By automating the cleaning of patterns and noise with regex, we aim to improve efficiency and reduce manual effort. The regular expressions are tuned iteratively based on continuously sampling and inspecting. This method allows us to programmatically clean a significant portion of repetitive issues in parsed PDF data before manual review. In the future, we can continue refining the regex and identifying new patterns to clean the data. The goal is to minimize the amount of manual effort while ensuring high-quality cleaned data. Some utilized code of regular expression (Python) can be seen as follows:
\begin{lstlisting}
r'\.{2,}'
r'\{2,}'
r'\s\s+'
r'(\d{1,3}\.){3}\d{1,3}|[a-f0-9]{32,64}'
r'^(\$.+[\n])+|^\w+\([^\)]+\)[\s\S]{1}\{[
\r]\s.+'
r'https?://(?:[-\w.]|(?:%[\da-fA-F]{2}))+'
\end{lstlisting}



\printcredits

\bibliographystyle{cas-model2-names}

\bibliography{cas-refs}


\bio{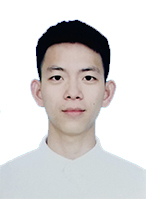}
{Wei Peng} received the B.S. degree in Computer Science and Technology from Chang’an University, Xi’an, China, in 2018 and his PhD degree in Institute of Information Engineering, Chinese Academy of Sciences, Beijing, China, in 2023. He has published	his research in high quality journals and conferences in the area, including KBs, Neurocomputing, IJCAI, SIGIR, AAAI, EMNLP,  etc. His research interests include summarization, natural language generation and question answering.
\endbio
\vspace{30pt}
\bio{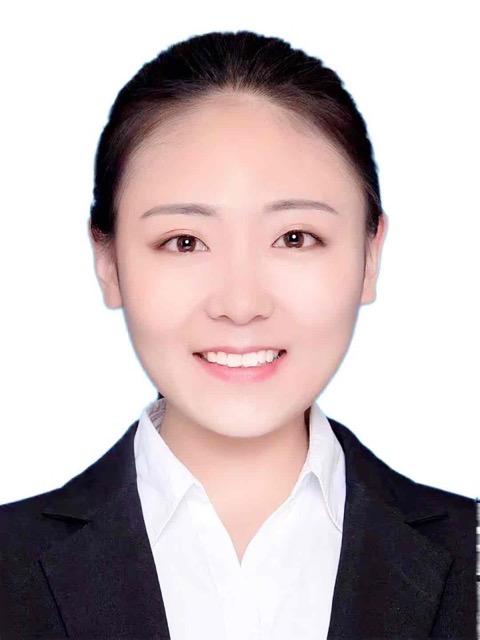}
{Junmei Ding} is a doctoral candidate at the School of Cyberspace Security, Beijing University of Posts and Telecommunications, holding a Master's degree in Software Engineering from Shanxi University. Her main research directions are anomaly behavior detection, threat detection, and cyber threat intelligence summarization.
\endbio
\vspace{30pt}
\bio{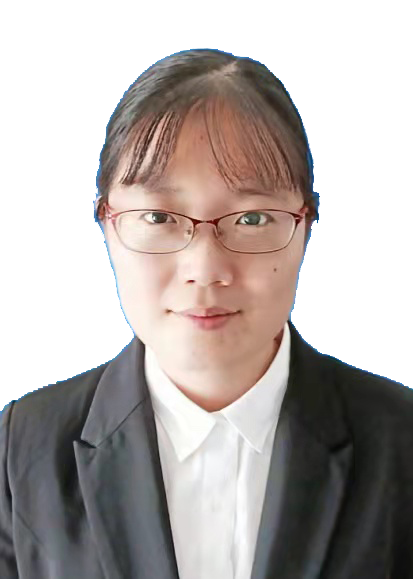}
{Wei Wang} received the doctor's degree in Institute of Information Engineering, School of Cyber Security, University of Chinese Academy of Sciences.  She is currently a research assistant of Zhongguancun Laboratory. Her research interests include system virtualization and network security. She has published several papers in journals and conferences including TSC, ICCD, VEE.
\endbio

\newpage
\bio{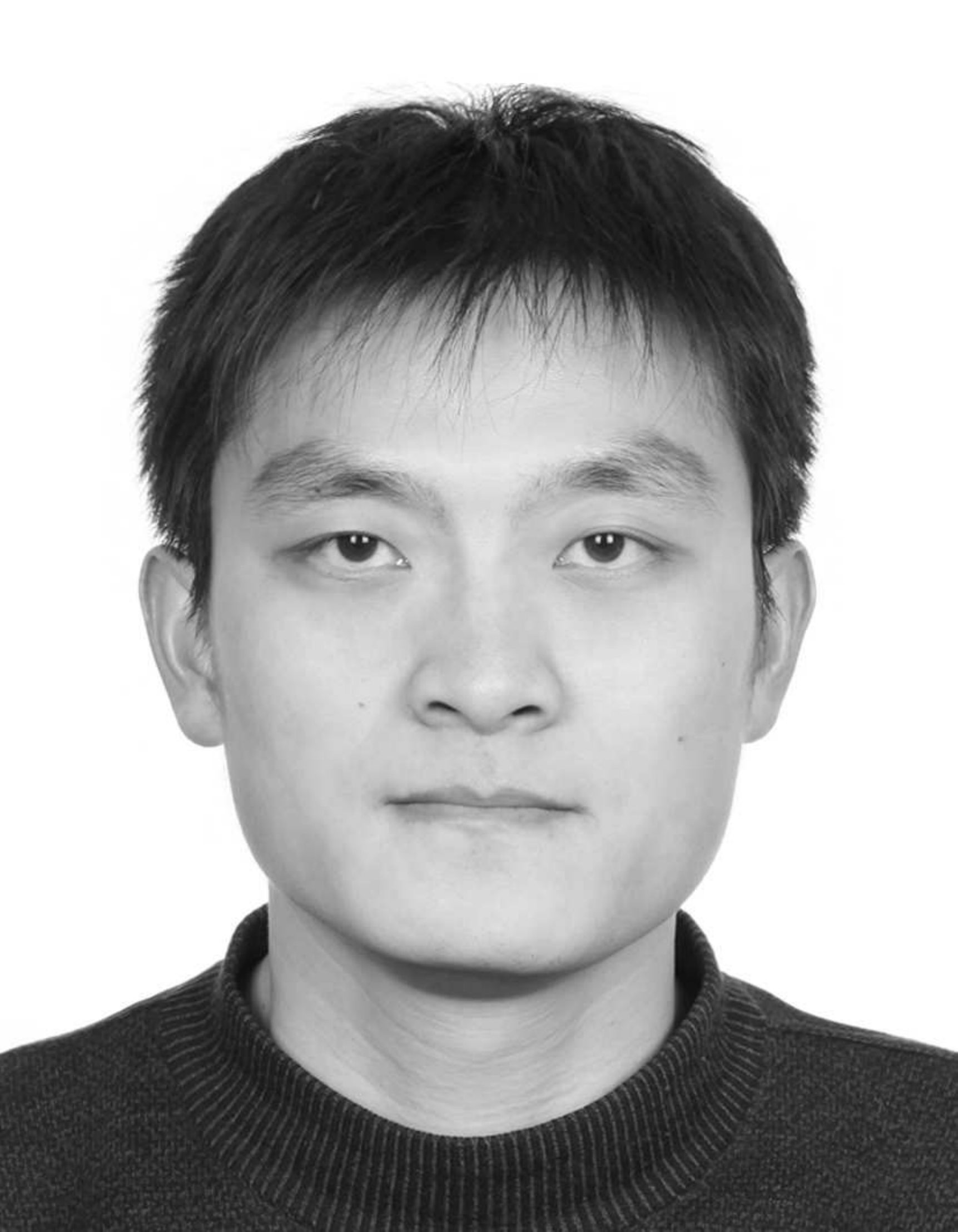}
{Lei Cui}
is an associate professor of Zhongguancun Laboratory, Beijing. He received his Doctor's degree in Computer Software and Theory from Beihang
University in 2015. His research interests include operating system, system security, and system virtualization. He has published over 40 papers in journals and conferences including TPDS, TIFS, TSC, ISSTA, ICCD, RAID, VEE, LISA, DSN.
\endbio

\vspace{30pt}
\bio{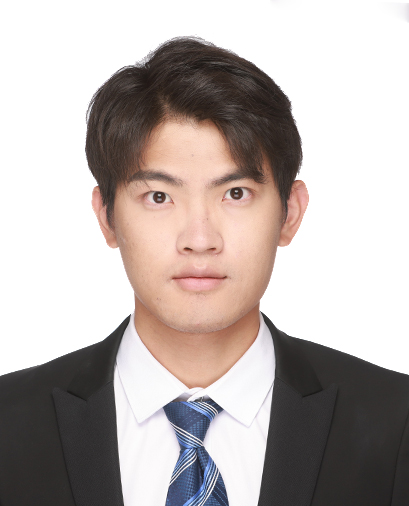}
{Wei Cai} is an assistant research scientist in the Network Connection Security Department at Zhongguancun Laboratory. He holds a PhD from the Institute of Information Engineering at the Chinese Academy of Sciences. His research primarily focuses on mobile encrypted traffic analysis and adversarial machine learning.
\endbio

\vspace{30pt}
\bio{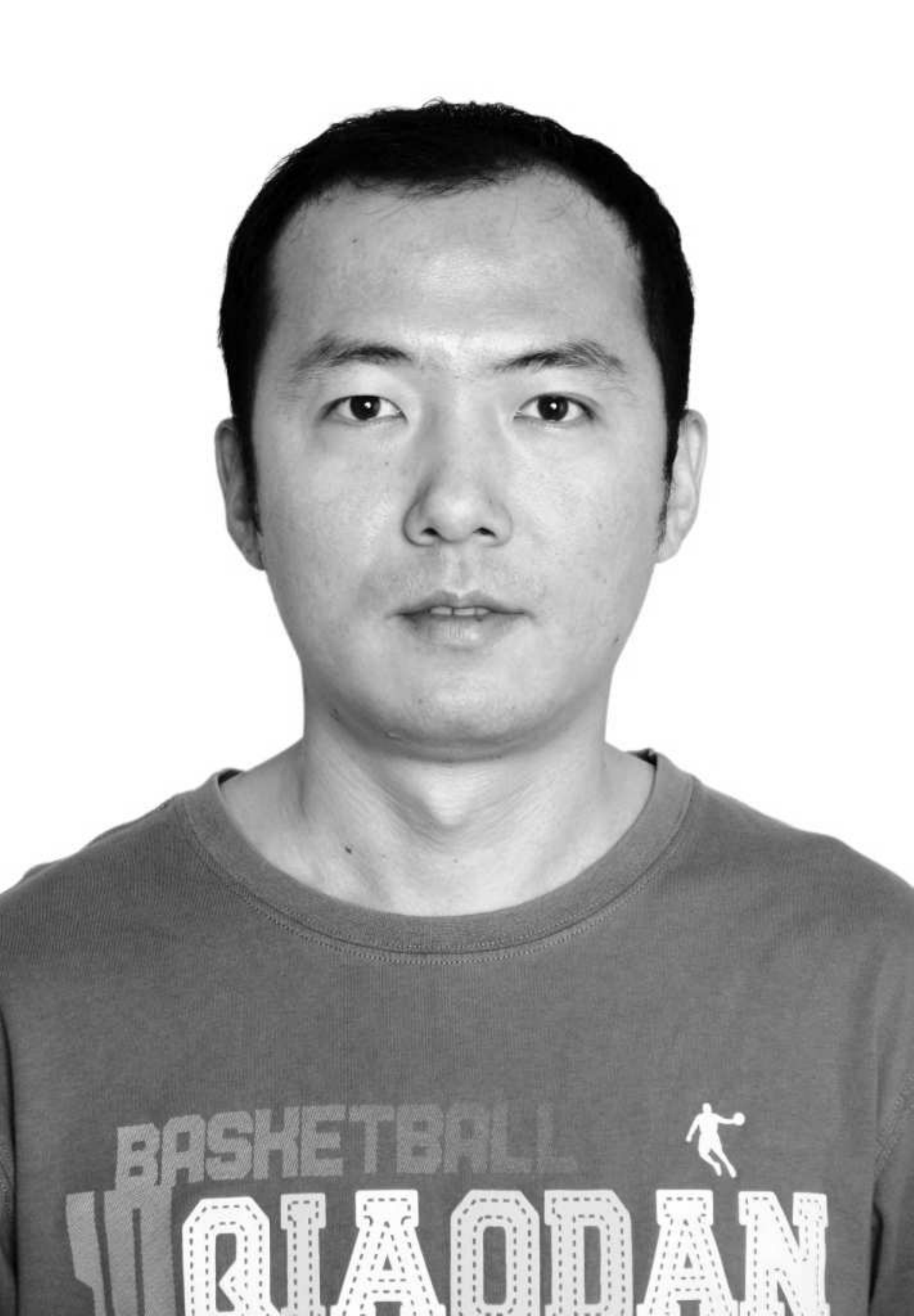}
{Zhiyu Hao}
is currently a professor of Zhongguancun Laboratory, Beijing. He received his Ph.D degree in Computer System Architecture from Harbin Institute of Technology in 2007. His research interests include network security, system virtualization. He has published over 50 papers in journals and conferences including TPDS, ICPP, IEEE S\&P, ICA3PP and CLUSTER.
\endbio

\vspace{30pt}

\bio{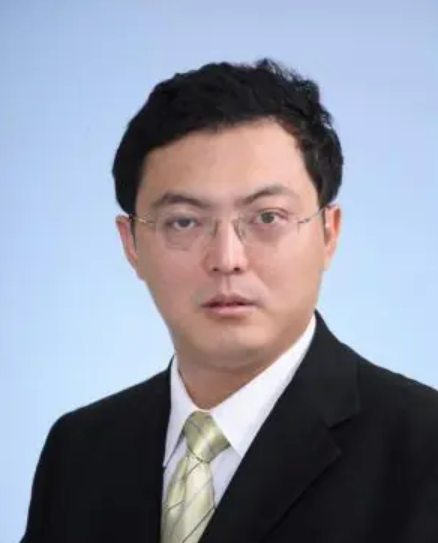}
{Xiaochun Yun} is the Deputy Director of the National Computer Network Emergency Response Technical Team/Coordination Center of China. He obtained a Bachelor's degree in Computer Science and Application from Harbin Institute of Technology, and a Ph.D. degree in Computer System Architecture from the same university. His research areas include network malware detection and prevention technology, security analysis, and content security. He has published over 100 academic papers.
\endbio

\end{document}